\documentclass{article}

\usepackage{microtype}
\usepackage{graphicx}

\usepackage{hyperref}

\usepackage{url}
\usepackage{booktabs}
\usepackage{multirow}

\usepackage{amsmath,amsfonts,amssymb,amsthm,color}
\usepackage{nicefrac}
\usepackage{xcolor}
\usepackage{algorithm,algorithmic}
\usepackage{caption}
\usepackage{graphicx}

\usepackage{soul}
\usepackage[normalem]{ulem}  %

\usepackage{subfig}
\usepackage{wrapfig}  %

\makeatletter
\newtheorem*{rep@theorem}{\rep@title}
\newcommand{\newreptheorem}[2]{%
\newenvironment{rep#1}[1]{%
 \def\rep@title{#2 \ref{##1}}%
 \begin{rep@theorem}}%
 {\end{rep@theorem}}}
\makeatother

\newreptheorem{theorem}{Theorem}
\newreptheorem{lemma}{Lemma}
\newreptheorem{definition}{Definition}
\newreptheorem{claim}{Claim}

\theoremstyle{definition}
\newtheorem{definition}{Definition}

\newcommand{\set}[1]{\mathcal{#1}}

\newcommand{\expectation}{\mathbb{E}}

\newcommand{\stdev}{$\pm$}

\def\eps{{\epsilon}}

\newcommand{\fcl}{\text{FCL}}
\newcommand{\similarity}{\mathrm{sim}}
\newcommand{\distance}{\mathrm{dist}}

\newcommand{\topk}{\textsc{Top}_k}
\newcommand{\bottomk}{\textsc{Bottom}_k}

\DeclareMathOperator*{\argmin}{arg\,min}

\usepackage[accepted]{icml2021arxiv}

\icmltitlerunning{Balancing Robustness and Sensitivity using Feature Contrastive Learning}

\begin{document}

\twocolumn[
\icmltitle{Balancing Robustness and Sensitivity using Feature Contrastive Learning}

\icmlsetsymbol{equal}{*}

\begin{icmlauthorlist}
\icmlauthor{Seungyeon Kim}{google}
\icmlauthor{Daniel Glasner}{google}
\icmlauthor{Srikumar Ramalingam}{google}
\icmlauthor{Cho-Jui Hsieh}{ucla}
\icmlauthor{Kishore Papineni}{google}
\icmlauthor{Sanjiv Kumar}{google}
\end{icmlauthorlist}

\icmlaffiliation{google}{Google Research}
\icmlaffiliation{ucla}{University of California, Los Angeles}
\icmlcorrespondingauthor{Seungyeon Kim}{seungyeonk@google.com}

\icmlkeywords{Machine Learning, ICML}

\vskip 0.3in
]

\printAffiliationsAndNotice{}  %


\begin{abstract}
It is generally believed that robust training of extremely large networks is critical to their success in real-world applications. However, when taken to the extreme, methods that promote robustness can hurt the model’s sensitivity to rare or underrepresented patterns. In this paper, we discuss this trade-off between sensitivity and robustness to natural (non-adversarial) perturbations by introducing two notions: \emph{contextual feature utility} and \emph{contextual feature sensitivity}. We propose Feature Contrastive Learning (FCL) that encourages a model to be more sensitive to the features that have higher contextual utility. Empirical results demonstrate that models trained with FCL achieve a better balance of robustness and sensitivity, leading to improved generalization in the presence of noise on both vision and NLP datasets.
\end{abstract}

\section{Introduction}

Deep learning has shown unprecedented success in numerous domains~\citep{Krizhevsky2012,Szegedy2015,He2016DeepRL,Hinton2012,sutskever2014sequence,Devlin2018}, and robustness plays a key role in the success of neural networks. When we seek robustness, as a general property of a model, we would like the model prediction to not change for small perturbations of the inputs. However, such invariance to small perturbations can be detrimental in some cases. As an extreme example, a small perturbation to the input can change the human perceived class label, but the model is insensitive to this change~\citep{tramer2020fundamental}. In this paper, we focus on balancing this trade-off between general robustness and sensitivity by developing a contrastive learning method. Contrasive learning is commonly used to learn visual representations~\citep{chen2020simple,He2020,Wu2018,Tian2020,khosla2020supervised}. Our goal is to promote change in model prediction for certain perturbations, and inhibit the change for the other perturbations. In this work we only address robustness to \emph{natural (non-adversarial) perturbations}. We do not attempt to improve robustness to carefully designed \emph{adversarial perturbations}~\citep{goodfellow2014explaining}. %

To develop algorithms that balance robustness and sensitivity, we first formalize two measures: \emph{utility} and \emph{sensitivity}. Utility refers to the change in the loss function when we perturb a specific input feature. Thus, a feature's utility is related to the model's prediction as well as the true label. Sensitivity, on the other hand, is the change in the learned embedding representation (before computing the loss) when we perturb a specific input feature. In contrast to classical feature selection approaches~\citep{guyon2003introduction,yu2004efficient} that identify relevant and important features, our notions of sensitivity and utility are \emph{context dependent} and change from one input to another. Our goal is to learn a model that is sensitive to high-utility features while still being robust to the perturbations of low-utility features.

\begin{figure*}[ht]
  \vspace{-1em}
  \centering
  \subfloat[\footnotesize Classes 0-4]{
        \includegraphics[width=0.32\textwidth]{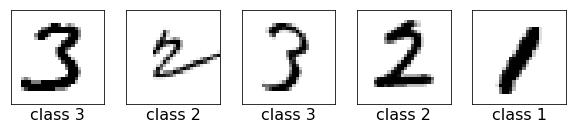}
        \label{fig:synthetic_mnist_0_to_4}
  }
  \subfloat[\footnotesize Classes 5-9]{
        \includegraphics[width=0.32\textwidth]{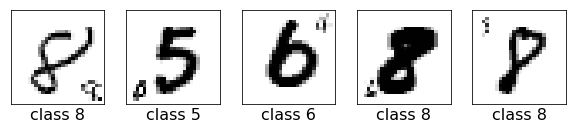}
        \label{fig:synthetic_mnist_5_to_9}
  }
  \subfloat[\footnotesize Classes 10-14]{
        \includegraphics[width=0.32\textwidth]{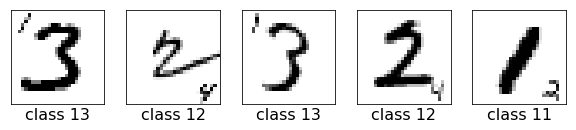}
      \label{fig:synthetic_mnist_10_to_14}
  }
  \vspace{-0.2em}
  \caption{Synthetic MNIST data. We synthesize new images by adding a scaled down version of a random digit to a random corner. Images synthesized from digits 5-9 keep their label (Figure \ref{fig:synthetic_mnist_5_to_9}) while images synthesized from digits 0-4 are considered to be of a different class (Figure {\ref{fig:synthetic_mnist_10_to_14}). In this setup corner pixels are informative only in a certain context.}}
  \label{fig:synthetic_mnist}
\end{figure*}

To explore and illustrate the notions of utility and sensitivity, we introduce a synthetic MNIST dataset, as shown in Figure~\ref{fig:synthetic_mnist}. In the standard MNIST, the goal is to classify 10 digits based on their appearance. We modify the data by adding a small random digit in the corner of some of the images and increasing the number of classes by five. For digits 5-9 we never change the class labels even in the presence of a corner digit, whereas digits 0-4 move to extended class labels 10-14 in the presence of any corner digit. The small corner digits can have high or low utility \emph{depending on the context}. If the digit in the center is in 5-9 the corner digit has no bearing on the class, and will have low utility. However, if the digit in the center of the image is in 0-4, the presence of a corner digit is essential to determining the label, and thus has high utility. 
We would like to promote model sensitivity to the small corner digits when they are informative, in order to improve predictions, but demote it when they are not, in order to improve robustness.

\paragraph{Feature attribution methods.}
Our notions of utility and sensitivity are related to feature attribution methods. Given an instance $x$ and a model $f$, feature based explanation aims to attribute the prediction of $f(x)$ to each feature. There have been two principal approaches to understand the role of features. In the first, we compute the derivative of $f(x)$ with respect to each feature, which is similar to the sensitivity measure proposed in this paper~\citep{shrikumar2017learning,smilkov2017smoothgrad,simonyan2013deep,SundararajanTY16}. %
The second approach measures the importance of a feature by removing it or comparing it with a reference point~\citep{samek2016evaluating,Fong2017Interpretable, Dabkowski2017Real,ancona2017unified,yeh2019on,Zeiler14visualizingand,zintgraf2017visualizing}. For example, the idea of prediction difference analysis is to study the regions in the input image that provide the best evidence for a specific class (or object) by studying how the prediction changes in the absence of a specific feature. 
While many of the existing methods look at the interpretability of the model predictions, our work proposes a new loss function in the training stage to adjust the sensitivity according to their utility in a context-dependent manner.

\paragraph{Robustness to natural perturbation vs. adversarial perturbation.}
It is widely believed that imposing robustness constraints or regularization to neural networks can improve their performance. Taking the idea of robustness to the extreme, adversarial training algorithms aim to make neural networks robust to any perturbation within an $\epsilon$-ball~\citep{goodfellow2014explaining, madry2017towards}. The certified defense methods pose an even stronger constraint in training, i.e., the improved robustness has to be verifiable~\citep{wong2018provable,zhang2019towards}. Despite being successful in boosting accuracy under adversarial attacks, they come at the cost of significantly degrading clean accuracy ~\citep{madry2017towards,zhang2019theoretically,wang2019bilateral}. Several theoretical works have demonstrated that a trade-off between adversarial robustness and generalization exists~\citep{tsipras2018robustness,schmidt2018adversarially}. Recent papers \citep{laugros2019adversarial, gulshad2020adversarial} also discuss the particular relationship between adversarial robustness and natural perturbation robustness, and find that they are usually poorly correlated. For example, \citet{laugros2019adversarial} shows that models trained for adversarial robustness are not more robust than standard models on common perturbation benchmarks and that the converse holds as well. \citet{gulshad2020adversarial} also found that natural robustness can commonly improve adversarial robustness slightly.
While adversarial robustness is important in its own way, this paper focuses on natural perturbation robustness. In fact, our goal of “making models sensitive to important features” implies that \emph{the model should not be adversarially robust on high utility features}.

With the goal of improving generalization instead of adversarial robustness, several other works enforce a weaker notion of robustness. A simple approach is to add Gaussian noise to the input features in the training phase. \citet{lopes2019improving} recently showed that Gaussian data augmentation with randomly chosen patches can improve generalization. \citet{xie2020adversarial} showed that adversarial training with a dual batch normalization approach can improve the performance of neural networks.

\paragraph{Contrastive learning for robustness}
It is worth noting that~\citep{kim2020adversarial,Jiang2020} also employ contrastive learning for robustness (see Section~\ref{sec:fcl} for details). However, our work is fundamentally different since our goal is to improve the model by contrasting at the feature level (high utility features vs. low utility features), while previous works contrast between different samples. Moreover, a) they focus on adversarial robustness while we focus on robustness to natural perturbations, b) their contrastive learning always suppresses the distance between the original and an adversarially perturbed input while ours increases the distance between high-utility perturbation pairs (\emph{this is the opposite direction of adversarial robustness}) and suppresses the distance for low-utility pairs, c) their perturbation is based on an unsupervised loss, while we rely on class labels to identify low and high utility features with respect to the classification task.

In summary, all the previous works in robust training aim to make the model insensitive to perturbation, \emph{while we argue that a good model (with better generalization performance) should be robust to unimportant features while being sensitive to important features.} A recent paper in the adversarial robustness community also pointed out this issue~\citep{tramer2020fundamental}, where they showed that existing adversarial training methods tend to make models {\it overly robust} to certain perturbations that the models should be sensitive to. However, they did not provide any solution to the problem of balancing robustness and sensitivity.

\paragraph{Main contributions}
\begin{itemize}
    \item We propose \emph{contextual sensitivity} and \emph{contextual utility} concepts that allow to measure and identify high utility features and their associated sensitivity (\S\ref{sec:formulation}).
    \item We propose Feature Contrastive Learning (FCL) that promotes model sensitivity to perturbations of high utility features, and inhibits model sensitivity to perturbations of low utility features (\S\ref{sec:fcl}).
    \item Using a human-annotated dataset, we verify that FCL indeed promotes sensitivity to high-utility features (\S\ref{sec:exp-sentiment}), and demonstrate its benefits on a noisy synthetic dataset (\S\ref{sec:exp-synthetic}) and on real-world datasets with corruptions (\S\ref{sec:exp-large}).
\end{itemize}

\section{Robustness and Sensitivity}\label{sec:formulation}

\subsection{Background and notation}

Before formally defining contextual utility and contextual sensitivity, we discuss a motivating example. Consider a sentence classification task with 0/1 loss. For a given sentence, removing one word can change the model’s prediction (i.e. its best guess). Removing the word can also change the true label. When the prediction changes, we say that the model is \emph{contextually} sensitive to this word. Note that sensitivity is independent of the true label. Contextual utility, on the other hand, is defined using the loss, which depends on the prediction as well as the true label. Even if the prediction changes, the loss may or may not be affected by the change because the true label can also change.

While the two concepts, utility and sensitivity, are related, neither implies the other. On the one hand, when both the prediction and the true label change, the 0/1 loss does not change; hence, the model is sensitive to the word, but the word's utility is zero. On the other hand, when only the true label changes, the model is not sensitive to the word, but the word has high utility. Ideally, we would like the model to be sensitive to features that have high utility.

We can naturally generalize these concepts to multi-class classification and relate sensitivity to the model’s probability distribution over the classes - rather than focusing on its best guess. Sensitivity can also be naturally defined with respect to a change in the logits, or in the embedding representation at any given layer in a deep neural network. We highlight one choice in the formal definition below, and use it in all our experiments.

\paragraph{Multiclass classification}

Consider a classification setting with $L$ classes. We are given a finite set of $n$ training samples $\set{S} = \left\{ (x_1, y_1), \dots, (x_n,y_n) \right\}$, where $x_i\in \set{X}$ and $y_i \in \set{Y}$. Here $\set{X}$ and $\set{Y}$ denote the instance and output spaces with dimensions $D$ and $L$ respectively. The output vector $y_i$ is the 1-hot encoding of the class labels. Let $f: \set{X} \rightarrow \mathbb{R}^L $ be the function that maps the input vector to one of the $L$ classes. Accordingly, given a loss function $\ell: \{0,1\}^L \times \mathbb{R}^L \rightarrow \mathbb{R}_{+}$, our goal is to find the parameters $w^*$ that minimize the expected loss:
\begin{align*}
    w^* = \argmin_w \expectation_{y \sim \set{Y}, x \sim \set{X}}  \ell(y, f(x;w)) .
\end{align*}
In this work, we consider the cross entropy loss function $\ell(y, f_w(x)) = \sum_c \mathbf{1}_{y=c} \log f(x; w)_c$, but our formulation is not restricted to this loss. The model $f(x): \set{X} \rightarrow \mathbb{R}^L $ can be seen as the composition of an embedding function $\phi: \set{X} \rightarrow \mathbb{R}^E$ that maps an input to an $E$-dimensional feature, and a discriminator function $h:  \mathbb{R}^E \rightarrow \mathbb{R}^L$ that maps a learned embedding to an output. In other words, $f(x;w) = (h \circ \phi)(x;w_{\phi},w_h)$ and $w=\{w_{\phi},w_h\}$.

Given a finite training set $\set{S}$, we minimize the following empirical risk to learn the parameters:
\begin{align*}
    w^* = \argmin_w \frac{1}{n} \sum_{(x_i, y_i) \sim \set{S}} \ell \left( y_i, f(x_i; w) \right).
\end{align*}

\subsection{Contextual feature utility}\label{sec:utility}

\begin{definition}[\emph{Contextual feature utility}] \label{def:utility}
Given a model $f: \set{X} \rightarrow \mathbb{R}^L $ and a loss function $\ell: \{0,1\}^L \times \mathbb{R}^L \rightarrow \mathbb{R}_{+}$, the \emph{contextual utility vector}, $u_i$, associated with a training sample $(x_i,y_i) \in \set{S}$, is given by:
\begin{align}
  u_{ij} &= \left| \frac{ \partial \ell(y_i, f(x_i; w))}{\partial x_{ij}} \right|
\end{align}
 when $x$ is continuous, and by
\begin{align}
 u_{ij}= \left| \ell(y_i, f(x_i; w)) - \ell(y_i, f(x_i\setminus{x_{ij}}; w)) \right|
\end{align}
when $x$ is discrete. Here $i$ is the index of a training sample and $j$ is the index of a feature of $x_i$, and $x_i\setminus{x_{ij}}$ denotes the example  $x_i$ with the $j$th feature removed.

In the continuous case, note that the contextual feature utility vector is nothing but the absolute value of Jacobian of the loss function with respect to the input vector, and the Jacobian has been shown to be closely related to stability of the network~\citep{jakubovitz2018improving}. 
\end{definition}

The contextual utility $u_{ij}$ denotes the change in the loss function $\ell$ with respect to perturbation of the input sample $x_i$ along the dimension $j$. A perturbation of the high utility feature leads to a larger change in loss compared to the perturbation of the low utility feature. Please note that this utility function is context sensitive, i.e., a dimension with high utility for one training sample may have low utility for another sample.

\subsection{Contextual feature sensitivity}\label{sec:sensitivity}

\begin{definition}[\emph{Contextual feature sensitivity}] \label{def:sensitivity}
Given an embedding function $\phi: \set{X} \rightarrow \mathbb{R}^L$, the sensitivity $s_{ij}$ associated with a training sample $(x_i,y_i) \in \set{S}$ and a feature index $j$ is given by:
\begin{align}
 s_{ij} =  \left\| \frac{ \partial \phi(x_i,w_{\phi})}{\partial x_{ij}} \right\|
\end{align}
when $x$ is continuous, and by
\begin{align}
 s_{ij} =  \left\| \phi(x_i,w_{\phi}) - \phi(x_i \setminus x_{ij},w_{\phi}) \right\|
\end{align} when $x$ is discrete.
\end{definition}
Sensitivity is nothing but the norm of the Jacobian of the embedding function with respect to the input. The notion of sensitivity captures how the embedding corresponding to an input $x_i$ changes for small perturbations of the input along dimension $j$. Similar to utility, the sensitivity is also context dependent and changes from one training sample to another. Note that the sensitivity could also be defined on the embeddings from intermediate layers, as well as the final output space. Driven by the empirical success of other stability training~\citep{zheng2016improving} and contrastive learning methods~\citep{chen2020simple}, we choose to develop contrastive loss functions in the embedding space defined by the penultimate layer of the network. In contrast to the feature utility vector that depends on the true class labels, the feature sensitivity is independent of the class labels. Please see Appendix~\ref{appendix:connection} for a more detailed discussion of the relationship between contextual feature utility and sensitivity.

\section{Feature contrastive learning}\label{sec:fcl}

\begin{algorithm}[!t]
    \caption{FCL algorithm}
	\begin{algorithmic}
	    \STATE Initialize model $f: \set{X} \rightarrow \mathbb{R}^L $ with parameters $w_0$
	    \FOR {Sample minibatch $S=[(x_1,y_1),...(x_n,y_n)]$ from $\set{S}$}
    	    \STATE $\forall_{i} \; u_{i}= \left| \frac{ \partial \ell(y_i, f(x_i; w))}{\partial x_{i}} \right|, $
    	    \FOR {$i \in \{1, ... n\}$} 
    	        \STATE $z_i = \phi(x_i,w_{\phi})$
    	        \STATE $z_i^{+} = \phi(x_i + \epsilon(\bottomk(u_i)),w_{\phi})$
    	        \STATE $z_i^{-} = \phi(x_i + \epsilon(\topk(u_i)),w_{\phi})$
    	    \ENDFOR
    	    \STATE $\ell_{\fcl} =  \sum_{i} \ell_{\fcl}^i$
    	    \STATE Update model parameters: $w_{t+1} \leftarrow w_t - \eta {\frac{ \partial \ell + \lambda \ell_{\fcl}}{\partial w}}$.
		\ENDFOR
	\end{algorithmic}\label{algo-fcl}
\end{algorithm}

Our goal is to learn an embedding function $\phi: \set{X} \rightarrow \mathbb{R}^L$ that is more \emph{sensitive} to the features with higher contextual utility than the ones with lower contextual utility. That is, we want embeddings of examples perturbed along low utility dimensions to remain close to the original embeddings, and embeddings of examples perturbed along high utility dimensions to be far. Our formulation utilizes the contextual utility and sensitivity and the interplay between them. The utility is used for selecting the features, and the associated sensitivity values are adjusted by applying the contrastive loss.

We now describe a method to achieve this goal, using a contrastive loss on embeddings, derived from utility-aware perturbations. In typical contrastive learning methods~\citep{chen2020simple}, positive and negative pairs are generated using data augmentations of the inputs, and the contrastive loss function minimizes the distance between embeddings from positive pairs, and maximizes the distances between embeddings from negative pairs. We follow the same path, but use contextual utility to define the positive and negative sets.

\begin{definition}[\emph{Utility-aware perturbations}] Let $\topk(v)$ and $\bottomk(v)$ denote the largest and smallest $k$ indices of vector $v$ (ties resolved arbitrarily), respectively. Let $\epsilon(\set{S})$ denote perturbation vectors of dimension $D$ such that
\begin{equation}
    \epsilon(\set{S})_i
    \begin{cases}
      \sim \mathcal{N}(0,\,\sigma^{2}). & \text{if}\ i \in \set{S} \\
      = 0. & \text{otherwise}
    \end{cases}
\end{equation}
In the discrete case a perturbation is the removal of a particular feature (or a token in NLP settings). Using the utility vector $u_i$ for a training sample $x_i$, we refer to  $\epsilon(\topk(u_i))$ as the high-utility perturbation, and $\epsilon(\bottomk(u_i))$ as the low-utility perturbation.
\end{definition}

For simplicity, let us use $z=\phi(x,w_{\phi})$ to denote the embedding associated with the input $x$. In order to increase the sensitivity along high utility features, we add a high-utility perturbation, $z_i^{-}= \phi(x_i + \epsilon(\topk(u_i)),w_{\phi})$. Similarly, in order to decrease the sensitivity along low utility features, we add a low-utility perturbation, $z_i^{+}= \phi(x_i + \epsilon(\bottomk(u_i)),w_{\phi})$. Our key idea is to treat $(z_i,z_i^{+})$ as a positive pair, and $(z_i,z_i^{-})$ as a negative pair in a contrastive loss. In other words, we want to do deep metric learning such that the high-utility perturbations lead to distant points and low-utility perturbations lead to nearby points in the embedding space.

For a given sample $x_i$, we have a single positive pair $\mathcal{P}_i=\{(z_i,z_i^{+})\}$ and a set of negative pairs $\mathcal{N}_i$, which consists of $(z_i,z_i^{-})$ and $(z_i,z_j)$ where $j\ne i$. We can now adapt any contrastive loss from the literature to our positive and negative pairs. In this paper we focus on the following choice. (See Appendix~\ref{appendix:loss} for a discussion of an alternative.)

\begin{figure*}[t]
    \vspace{-0.2em}
    \centering
    \includegraphics[width=0.65\linewidth, trim=140 450 140 60, clip]{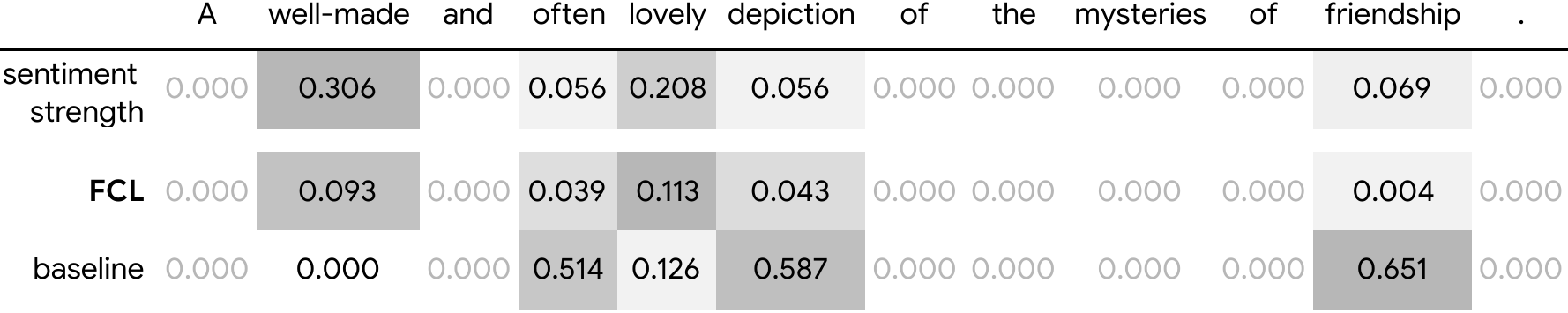}
    \vspace{-1em}
    \caption{An example sentence from the SST dataset demonstrating per-token human sentiment-strength annotations (how far each sentiment is from the neutral state). The two bottom rows show the sensitivity of a trained model with only classification loss (`baseline') or with an additional FCL loss (`FCL'). The background-color in each cell represents the relative magnitude of the value in each row. `FCL''s sensitivity orderings align with the ground-truth much better than the baseline. For example, `baseline' is not sensitive to `well-made' and more sensitive to `often' or `friendship' than to `lovely', which is the opposite of the ground-truth.}
    \vspace{-1em}
    \label{fig:sst-example}
\end{figure*}

\begin{definition}[\emph{Feature Contrastive Loss}] Given the positive pair $\mathcal{P}_i$ and the set of negative pairs $\mathcal{N}_i$ for a sample $x_i$, we define the Feature Contrastive Loss ($\ell_\fcl$) as follows:
\begin{align}
    \ell_{\fcl}^i & = 
    - \log \frac{e^ {\similarity(z_i,z_i^{+})/\tau}} 
    {e^ {\similarity(z_i,z_i^{+})/\tau} + \sum_{(z_i,z_j) \in \mathcal{N}_i} e^{\similarity(z_i,z_j)/\tau}}, \label{eq:fcl-xe}
\end{align}
where $\similarity(a,b) = \frac{a^T b}{|a||b|}$, and $\tau$ is a temperature parameter. Our definition is similar to the recent contrastive learning method~\citep{chen2020simple}. 
\end{definition}

Algorithm~\ref{algo-fcl} describes the details of FCL algorithm. It's important to note that during early stages of training, the utility is likely to fluctuate and be very noisy. Imposing sensitivity constraints based on the early stage utility can be detrimental. We therefore use a warm-up schedule. We keep $\lambda = 0$ until a certain number of training epochs and then switch it to a fixed positive value for the rest of the training.

\paragraph{Discussion}
We can also use external sources for the utility; these can come as a replacement, or in addition to the model’s utility. This could be particularly useful in distillation and domain adaptation settings. For example, when developing a model with limited training data, utility values from the teacher, or the source task can be beneficial.

\section{Experiments}

\subsection{Sentiment understanding}\label{sec:exp-sentiment}

In this section, we apply FCL in a real world task and show how we can discover high utility features and increase their sensitivity values accordingly. We chose Stanford Sentiment Treebank (SST) dataset \cite{socher2013recursive} since it provides both sentence-level human annotations as well as per-token ones (see an example in Figure~\ref{fig:sst-example}). We design our experiment as follows.

\paragraph{Training}
Models are trained with \emph{sentence-level} binary sentiment labels and do not have any access to per-token sentiment scores. This setup is commonly referred to as GLUE SST-2 \cite{wang2018glue}. Thus, token sensitivity of models trained in this setup is determined by optimizing for sentence-level \emph{binary} labels.

\paragraph{Evaluation}
Models are evaluated by comparing the sequence of token sensitivities to the ground-truth sentiment \emph{strength}. We compute token sensitivities using Definition~\ref{def:sensitivity} (\S\ref{sec:sensitivity}). We hypothesize that the stronger sentiment tokens will have higher utility values, and we expect the FCL to increase their sensitivity values compared to weaker sentiment tokens. We focus our evaluation on relative ordering of tokens which carry some sentiment and ignore neutral sentiment tokens, whose relative ordering is somewhat arbitrary. Specifically, we ignore tokens with sentiment values in the mid range $[0.45, 0.55]$ around the neutral value $0.5$.

\begin{figure*}[!t]
    \centering
    \includegraphics[width=0.7\textwidth]{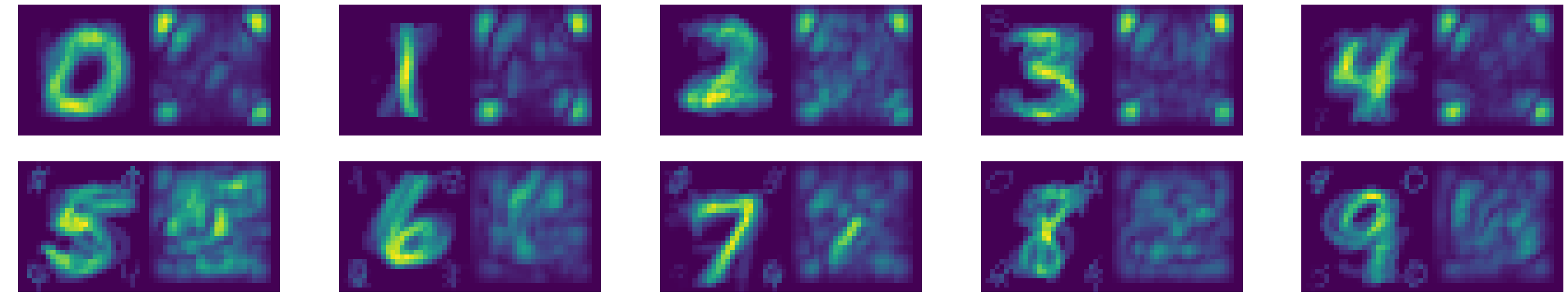}
    \vspace{-0.8em}
    \caption{Visualization of utility averaged over classes (modulo 10) within one batch. Each of the ten panels shows the average image on the left, and the average utility on the right. We can see that the corner pixels have high utility \emph{only in a certain context}. 
    When the central digit is 0-4, the corner pixels are important, since they can flip the class, but when the central digit is 5-9 they are not.}\label{fig:synthetic_mnist_vis}
\end{figure*}

\paragraph{Experiment} We compare a baseline trained with a cross-entropy classification loss for the binary sentiment task (`baseline') with a model trained with an additional FCL-loss~(Eq.\eqref{eq:fcl-xe}) on top of the cross-entropy loss (`FCL'). In both cases, we use a 3-stack Transformer~\cite{vaswani2017attention} with intermediate dimension 256 and 8 attention heads, followed by a linear classifier. Each sentence is processed with a BERT-tokenizer to generate token embeddings which are then fed to the transformer. We swept through $\tau=[1.0,...,0.02]$ and the fraction $[0.1, 0.2, 0.3]$ of the sequence length which we use as top/bottom $k$ features. We turn on the FCL-loss after 5k steps of training using only cross-entropy loss.

\paragraph{Result} Both `baseline` and `FCL` achieve the same high test split accuracy of $87.5\%$ for the binary classification task. Our accuracy matches the performance reported in \cite{wang2018glue} (c.f. $85.4\%$ accuracy reported in the original paper \cite{socher2013recursive}). However, the correlation of `baseline' and `FCL' with the per-token sentiment strength varies significantly. When using FCL-loss, the average Person correlation over the whole dataset increases significantly, from 0.6702 for the baseline to 0.7613 for FCL. In Figure~\ref{fig:sst-example} we show a qualitative comparison in which the per-token sensitivity from `FCL' aligns well with the ground truth while `baseline` does not.

\subsection{Synthetic MNIST classification}\label{sec:exp-synthetic}

Most datasets allow us to evaluate the robustness of an algorithm, and not the sensitivity. To illustrate how FCL can balance both robustness and sensitivity, we introduce a synthetic dataset based on MNIST digits \citep{lecun1989backpropagation}. We set up the task so that some patterns are not useful most of the time, but are very informative in a certain context, which occurs rarely. We show that by using FCL, our models i) maintain sensitivity \emph{in the right context} and ii) become more robust by suppressing uninformative features.

\begin{figure}[!t]
    \vspace{-1.5em}
    \begin{minipage}{\linewidth}
    \centering
    \subfloat[\footnotesize Train split (log scale)]{  %
        \includegraphics[width=0.47\textwidth]{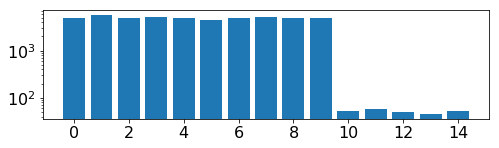}
        \label{fig:synthetic_mnist_train_dist}
    }
    \subfloat[\footnotesize Test split ]{  %
        \includegraphics[width=0.47\textwidth]{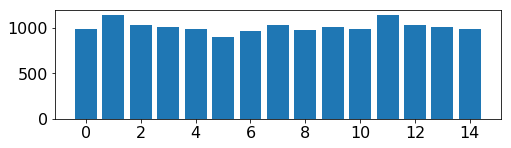}
        \label{fig:synthetic_mnist_test_dist}
    }
    \vspace{-0.5em}
    \caption{Class distribution. Train split is highly unbalanced with classes 10-14 appearing rarely.}
    \label{fig:synthetic_mnist_dist}
    \end{minipage}
    \vspace{0.5em}
    \begin{minipage}{\linewidth}
    \centering
    \subfloat[\footnotesize Uniform noise]{
        \includegraphics[width=0.47\textwidth]{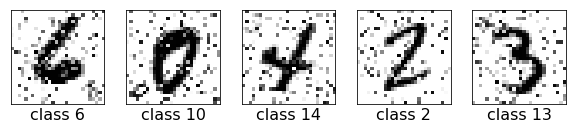}
        \label{fig:synthetic_mnist_uniform_noise}
    }
    \subfloat[\footnotesize Non-uniform noise]{
        \includegraphics[width=0.47\textwidth]{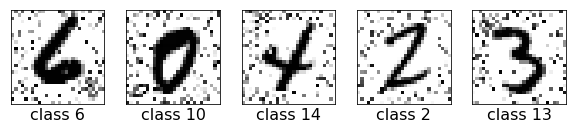}
        \label{fig:synthetic_mnist_std_dependent_noise}
    }
    \vspace{-0.5em}
    \caption{Two types of random noise, used to evaluate notions of robustness.}
    \label{fig:synthetic_mnist_noise}
    \end{minipage}
    \vspace{-1em}
\end{figure}

\paragraph{Data generation}
The original MNIST images consist of a single digit centered over a uniform background, the corners of the image are empty in almost all examples, as seen in Figure \ref{fig:synthetic_mnist_0_to_4}.
We synthesize new images by adding a scaled down version of a random digit to a random corner, as seen in Figures \ref{fig:synthetic_mnist_5_to_9} and \ref{fig:synthetic_mnist_10_to_14}.
The images synthesized from digits 0-4 are considered to be new classes, classes 10-14 respectively. Examples are shown in Figure \ref{fig:synthetic_mnist_10_to_14}.
In contrast, images synthesized from digits 5-9 do not change the class label, as shown in Figure \ref{fig:synthetic_mnist_5_to_9}.
For the new images, the small digits in the corners are uninformative except in a certain context. If the digit in the center is in 5-9 the corner digit has no bearing on the class, but if the digit in the center of the image is in 0-4, the presence of a corner digit is essential to determining if the image should be labeled as 0-4 or as 10-14.

\paragraph{Experiment}
We generate a training set in which the new classes, classes 10-14 are very rare (see Figure~\ref{fig:synthetic_mnist_train_dist}), appearing with a ratio of approximately $1/100$ compared to classes 0-9. Classes 0-9 have approximately $5000$ examples each, while classes 10-14 have approximately $50$ each. 
The challenge for models trained with this data is that the small digits in the corners are going to be completely uninformative 100 out of 101 times they appear.
To emphasize the importance of learning the rare classes, our test (and validation) sets have a balanced distribution over all classes (Figure~\ref{fig:synthetic_mnist_test_dist}). The balanced test set is labeled `BAL'. In total, we have roughly $50$k training examples, $15$k validation examples and $15$k test examples. The validation set has a distribution similar to the test set's and was used to tune hyper-parameters.

To demonstrate that FCL increases robustness to noise, we also prepare two noisy versions of the balanced test set. In both of these test sets we replace $15\%$ of the pixels, with a uniformly chosen random gray level. For the uniform noise test set (Figure \ref{fig:synthetic_mnist_uniform_noise}) the location of the noisy pixels is chosen uniformly. We label this set `BAL+UN'. For the \emph{non-uniform noise} test set, `BAL+NUN' (Figure \ref{fig:synthetic_mnist_std_dependent_noise}) the probability of a pixel being replaced with noise is inversely proportional to its sample standard deviation (over training images). The intuition is that in this set, noise will be concentrated in less ``informative'' pixels.

We train a LeNet-like convolutional neural network (CNN) ~\citep{lecun1989backpropagation}.
The network is trained for 20 epochs using the Adam optimizer~\citep{kingma2014adam}, with an initial learning rate of 0.01 and exponential decay at a rate of 0.89 per epoch. FCL is turned on after 2 epochs with a linear warmup of 2 epochs. We set $k=256, \lambda=0.001, \tau=0.1$ and $\sigma = 0.5$ (image values are in $[0, 1]$). These values were determined empirically using the validation set. In later stages of training the utility values become very small. To avoid numerical issues we drop high utility perturbations if the max utility value is smaller than $\eps=10^{-12}$. Each experiment is repeated 10 times.

\begin{table}[!t]
    \vspace{-0.5em}
    \footnotesize
    \centering
    \begin{tabular}{@{}lrrr@{}}
        \toprule
         & BAL & BAL+UN & BAL+NUN \\
        \midrule
        XE & 0.9250 \stdev0.0088 & 0.4123 \stdev0.1176  &  0.5473 \stdev0.0703\\
        \bf FCL & 0.9207 \stdev0.0129 & \bf 0.6384 \stdev0.0530 & \bf 0.6896 \stdev0.0349 \\
        \bottomrule
    \end{tabular}
    \vspace{-0.5em}
    \caption{Average accuracy over 10 runs on the synthetic MNIST data. Both methods are trained on the same unbalanced training set, and evaluated on balanced test sets: `BAL' (balanced), `BAL+UN' (balanced with added uniform noise) and `BAL+NUN' (balanced with added non-uniform noise). See text for details.}
    \label{table:synthetic_mnist_accuracy}
    \vspace{-0.5em}
\end{table}

\begin{table*}[t]
    \small
    \centering
    \begin{tabular}{@{}llrrr@{}}
        \toprule
        Dataset & Method & Clean & UN & NUN \\
        \midrule
        \multirow{5}{*}{\shortstack[l]{Noisy\\CIFAR-10}} 
            & XE 
            & 0.9389 \stdev0.0014 
            & 0.1317 \stdev0.0135 
            & 0.1256 \stdev0.0089 \\
            & XE+Gaussian 
            & 0.9375 \stdev0.0016
            & 0.3409 \stdev0.0580
            & 0.3175 \stdev0.0532\\
            & CL+Gaussian
            & 0.9362 \stdev0.0009
            & 0.2646 \stdev0.0165
            & 0.2464 \stdev0.0159\\
            & \bf FCL 
            & 0.9375 \stdev0.0010
            & \bf 0.3749 \stdev0.0293
            & \bf 0.3432 \stdev0.0231\\
            \cmidrule{2-5}
            & Patch Gaussian+XE
            & 0.9334 \stdev0.0035
            & 0.7842 \stdev0.0087
            & 0.7669 \stdev0.0086\\
            & Patch Gaussian+\textbf{FCL}
            & 0.9354 \stdev0.0023
            & \bf 0.8210 \stdev0.0013
            & \bf 0.8066 \stdev0.0033\\
        \midrule
        \multirow{5}{*}{\shortstack[l]{Noisy\\CIFAR-100}}
            & XE 
            & 0.7323 \stdev0.0052 
            & 0.0366 \stdev0.0078 
            & 0.0356 \stdev0.0084 \\
            & XE+Gaussian
            & 0.7297 \stdev0.0057
            & 0.0806 \stdev0.0187
            & 0.0763 \stdev0.0162\\ 
            & CL+Gaussian
            & 0.7294 \stdev0.0022
            & 0.0668 \stdev0.0122
            & 0.0640 \stdev0.0134\\
            & \bf FCL
            & 0.7252 \stdev0.0076
            & \bf 0.1477 \stdev0.0227
            & \bf 0.1007 \stdev0.0160\\
            \cmidrule{2-5}
            & Patch Gaussian+XE
            & 0.7315 \stdev0.0028
            & 0.0385 \stdev0.0102
            & 0.0377 \stdev0.0091\\
            & Patch Gaussian+\textbf{FCL}
            & 0.7254 \stdev0.0045
            & \bf 0.1590 \stdev0.0200
            & \bf 0.1033 \stdev0.0174\\
        \bottomrule
    \end{tabular}
    \vspace{-0.5em}
    \caption{Average accuracy and standard deviation on the noisy CIFAR test sets (5 runs). Methods which significantly outperform others in their group are highlighted with boldface. Please see Section~\ref{sec:exp-large} for descriptions of the baseline methods.}  \label{tbl:noisy-cifar}
\end{table*}

\paragraph{Results}
The mean accuracy and the standard deviation are shown in Table \ref{table:synthetic_mnist_accuracy}. Results on the noisy test sets `BAL+NU' and `BAL+NUN' show that using FCL can significantly improve robustness to noise while maintaining sensitivity. Figure~\ref{fig:synthetic_mnist_vis} illustrates the context dependent utility of the small digits in the corners of the image. This is the signal used by FCL to emphasize contextual sensitivity. Note that the models don't see any noisy images in training, they can however learn which pixels are less informative in certain contexts and suppress reliance on those.

\subsection{Larger-scale experiments}\label{sec:exp-large}

\begin{table*}[!t]
    \small
    \centering
    \begin{tabular}{@{}llrrrrrrr@{}}
        \toprule
        Dataset & Method & All average &  Noise & Blur & Weather & Digital \\
        \toprule
        CIFAR-10-C
        & XE 
        & 0.7137 \stdev0.0038
        & 0.4967 & 0.6833 & 0.8309 & 0.7537\\
        & XE+Gaussian
        & 0.7379 \stdev0.0089
        & 0.5800 & \bf 0.6967 & \bf 0.8389 & 0.7572\\
        & CL+Gaussian
        & 0.7253 \stdev0.0028
        & 0.5446 & 0.6939 & 0.8312 & 0.7530 \\
        & \bf FCL
        & \bf 0.7446 \stdev0.0055
        & \bf 0.6416 & 0.6886 & 0.8338 & 0.7530\\
        \cmidrule{2-7}
        & Patch Gaussian+XE
        & 0.8311 \stdev0.0027
        & 0.8951 & 0.7625 & 0.8540 & 0.8021\\
        & Patch Gaussian+\textbf{FCL}
        & 0.8319 \stdev0.0029
        & \bf 0.8993 & \bf 0.7639 & 0.8536 & 0.8000\\
        \toprule
        CIFAR-100-C 
        & XE
        & 0.4428 \stdev0.0038 
        & 0.2113 & 0.4323 & 0.5527 & 0.4855\\
        & XE+Gaussian
        & 0.4512 \stdev0.0067
        & 0.2502 & 0.4308 & 0.5524 & 0.4848\\
        & CL+Gaussian
        & 0.4480 \stdev0.0057
        & 0.2350 & 0.4350 & 0.5514 & 0.4865\\
        & \bf FCL
        & \bf 0.4706 \stdev0.0031
        & \bf 0.3528 & 0.4355 & 0.5467 & 0.4847\\
        \cmidrule{2-7}
        & Patch Gaussian+XE
        & 0.4448 \stdev0.0030
        & 0.2198 & 0.4344 & 0.5483 & 0.4896\\
        & Patch Gaussian+\textbf{FCL}
        & \bf 0.4742 \stdev0.0054
        & \bf 0.3699 & 0.4353 & 0.5490 & 0.4851\\
        \toprule
        ImageNet-C
        & XE 
        & 0.3406 \stdev0.0007
        & 0.2615 & 0.2816 & 0.4214 & 0.3783\\
        & XE+Gaussian
        & 0.3414 \stdev0.0012
        & 0.2623 & 0.2829 & \bf 0.4224 & 0.3783\\
        & CL+Gaussian
        & 0.3418 \stdev0.0016
        & 0.2658 & 0.2824 & 0.4223 & 0.3778\\
        & \bf FCL
        & \bf 0.3437 \stdev0.0022
        & \bf 0.2696 & 0.2850 & 0.4188 & \bf 0.3827\\
        \cmidrule{2-7}
        & Patch Gaussian+XE 
        & 0.3625 \stdev0.0023
        & 0.3053 & \bf 0.3041 & 0.4300 & 0.3964\\
        & Patch Gaussian+\textbf{FCL}
        & \bf 0.3634 \stdev0.0045
        & \bf 0.3077 & 0.3034 & 0.4308 & \bf 0.3976\\
        \bottomrule
    \end{tabular}
    \vspace{-0.3em}
    \caption{Image classification accuracy on the corrupted CIFAR and ImageNet datasets \citep{hendrycks2019benchmarking}. Results which are significantly better than others in their group are highlighted with boldface. The `All average' column summarizes performance on all 19 corruption patterns. The other columns show averages within each corruption group. The full table can be found in Appendix~\ref{appendix:full_corruptions_accuracy}.}
    \label{tbl:cifarc}
\end{table*}

To evaluate FCL's performance on general tasks, we conducted experiments on public large-scale image datasets (CIFAR-10, CIFAR-100, ImageNet) with synthetic noise injection similar to Section~\ref{sec:exp-synthetic}, and with the 19 predefined corruption patterns from \citep{hendrycks2019benchmarking} -- called CIFAR-10-C, CIFAR-100-C and ImageNet-C. We show that FCL can significantly improve robustness to these noise patterns, with minimal, if any, sacrifice in accuracy.

\paragraph{Baselines}
Apart from the standard cross-entropy baseline `XE', we consider three other baselines `XE+Gaussian', `CL+Gaussian' and `Patch Gaussian+XE'. In `XE+Gaussian', all the image pixels are perturbed by Gaussian noise, and an additional cross-entropy term (weighted by a scalar $\lambda$) is applied to perturbed versions of the image, keeping the original label. In `CL+Gaussian', we add a contrastive loss similar in form to $\ell_{\fcl}$~\eqref{eq:fcl-xe} to the original cross-entropy classification loss. We use the same weight $\lambda$ as in FCL but with a random Gaussian perturbed image as the positive pair instead of the utility-dependent perturbation. `Patch Gaussian', recently proposed by~\citep{lopes2019improving} is a data augmentation technique. An augmentation is generated by adding a patch of random Gaussian noise to a random position in the image. This technique achieved state-of-the-art performance on CIFAR-10-C.
In `XE+Gaussian' the perturbation is applied to all features, in `Patch Gaussian+XE' it is applied to a subset of the pixels, chosen at random, while FCL applies perturbations to a subset of pixels based on \emph{contextual utility}. Note that since Patch Gaussian is purely a data augmentation technique, it can easily be combined with FCL, as we do in `Patch Gaussian+FCL'.

\paragraph{Model and hyperparameters} ResNet-56 was used for CIFAR experiments and ResNet-v2-50 for the ImageNet experiment. We used the same common hyper-parameters such as learning rate schedule and the use of SGD momentum optimizer ($0.9$ momentum) across all experiments. Details on hyper-parameters,  learning rate schedules and optimization can be found in Appendix~\ref{appendix:exp}. Models are trained for 450 epochs and contrastive learning losses (FCL and CL+Gaussian) are applied after 300 epochs (CIFAR) or 60 epochs (ImageNet). We kept all standard CIFAR/ImageNet data augmentations (random cropping and flipping) across all runs and added Patch Gaussian before or after the standard data augmentation as in \citep{lopes2019improving} when specified. For both Gaussian noise baselines, we swept $\sigma=[0.1, 0.3, 0.5]$ to choose the best performing parameter. For the Patch Gaussian, we used the code and the recommended configurations from \citep{lopes2019improving} -- CIFAR-10: patch size=$25$, $\sigma=0.1$, ImageNet: patch size$\le250$, $\sigma=1.0$. Since CIFAR-100 parameters were not provided from the paper, we started from CIFAR-10 parameters and made our best effort to sweep the parameters (patch size=$[15...25]$, $\sigma=[0.01...0.1])$. For contrastive learning methods, we swept $\lambda=[0.0001...0.0004]$ and $\tau=[2, 1, 0.5, 0.1]$. For FCL, we swept $k=[256,512,1024,2048]$ and $\sigma_\epsilon=[0.1, 0.3, 0.5]$. We repeated all experiments $5$ times.

\subsubsection{Noisy CIFAR images}

We follow the same protocol described in the synthetic MNIST experiment (Section~\ref{sec:exp-synthetic}) to generate uniform noise `UN' and non-uniform noise `NUN' test sets for CIFAR-10 and CIFAR-100. Table~\ref{tbl:noisy-cifar} demonstrates that FCL outperforms all baseline models with or without the PG data augmentation. 
\paragraph{Noisy CIFAR-10} We can observe that Gaussian perturbation does improve performance in both UN and NUN (XE vs. XE+Gaussian or XE vs. CL+Gaussian); however, FCL's selective perturbation on the high contextual utility features obtains a better improvement in all cases (both Gaussian baselines vs. FCL). When combined with the PG data augmentation, the gap between the clean accuracy versus `UN' or `NUN' narrows (0.93 vs 0.82). The combined version (Patch Gaussian+FCL) achieves the best noisy CIFAR-10 performance (on `UN' and `NUN'), without hurting the clean accuracy .

\paragraph{Noisy CIFAR-100} Without the PG data augmentation, the pattern is similar to the case above; however gaps between XE, Gaussian baselines and FCL are wider suggesting that FCL gives more benefit when the number of classes is larger. PG did not work well in the 100 class setting, even with extensive tuning (including the recommended configurations from \citep{lopes2019improving}). The combination (Patch Gaussian+FCL) achieves the best performance on `UN' and `NUN'.

\subsubsection{CIFAR-10-C, CIFAR-100-C and ImageNet-C}

We conducted a similar experiment on the public benchmark set of corrupted images \citep{hendrycks2019benchmarking}. This benchmark set evaluates robustness to natural perturbations of a prediction model by applying 19 common corruption patterns to CIFAR and ImageNet images. Table~\ref{tbl:cifarc} shows the averaged accuracy on  all corruption patterns, as well as the averages from each corruption pattern group. The full results are provided in Appendix~\ref{appendix:full_corruptions_accuracy}.

\paragraph{CIFAR-10-C} The pattern is similar to the noisy CIFAR-10. Adding the Gaussian perturbation improves the average performance, particularly in the noise-corruption pattern group and the blurring group. FCL works much better than Gaussian baselines providing an additional large improvement in the noise group. PG augmentation works really well for this task, particularly in the noise and blur groups (currently it is a state-of-the-art). Nevertheless, adding FCL can still add some value to some of the patterns. Appendix~\ref{appendix:full_corruptions_accuracy} shows FCL performs better for impulse noise and zoom blur patterns, while the vanilla PG performs better in fog and pixelization.

\paragraph{CIFAR-100-C} Without PG, the improvement of FCL over the other baselines is even larger than for CIFAR-10-C, with a drastic improvement on the noisy corruption group. Similar to Noisy CIFAR-100, PG did not perform well in this setting, while PG+FCL still was able to perform well, achieving even better accuracy than without PG.

\paragraph{ImageNet-C}
With or without PG, FCL outperforms the baselines with the large improvements on `digital', and `noise' corruption patterns. Among individual patterns (reported in Appendix~\ref{appendix:full_corruptions_accuracy}), FCL performs particularly well on the `shot' corruption pattern.

\section{Summary}

In this paper, we propose Feature Contrastive Learning (FCL), a novel approach to balance robustness and sensitivity in deep neural network training. While most prior work focus on always increasing the robustness and decreasing the sensitivity to feature perturbations, we argue that it is important to strike a balance and selectively enhance robustness and sensitivity based on the context.

\bibliography{09-references}
\bibliographystyle{icml2021}

\clearpage
\appendix
\onecolumn
{\Large\bf Supplementary material for ``Balancing Robustness and Sensitivity using Feature Contrastive Learning''}

\section{Connection between contextual feature utility and sensitivity}\label{appendix:connection}

Consider a classification task with cross entropy loss, and let $f()$ be the output of the network after applying a softmax. In this setting, the loss is minus log probability of the correct label.
The contextual utility of $f()$ for a given feature is defined by 
\begin{align}
  u = \left| \frac{ \partial \ell(y, f(x; w)) }{ \partial x} \right| = \left| \frac{ \partial \log[ f(x; w)_{y} ] } { \partial x } \right| = \frac{1}{f(x; w)_y} \left| \frac{\partial f(x; w)_y }{ \partial x } \right|.
\end{align}
Also recall that the contextual sensitivity of $f()$ for a given feature is given by
\begin{align}
    s = \left\| \frac{\partial f(x; w)}{\partial x} \right\| = \sqrt{\frac{\partial f(x; w)_y^2}{\partial x} + \sum_{c \neq y} \frac{\partial f(x; w)_c^2}{\partial x}}.
\end{align}

We can see that the contextual feature utility is a product of two terms. The first is the reciprocal of the networks’ prediction for the correct class, and the second is a sensitivity-like term specific to the correct class. When the network’s prediction is correct the utility is proportional to the ground truth class’s sensitivity. If changing the feature will not affect the correct prediction it doesn’t have much utility and vice versa. On the other hand, when the network makes a mistake, the utility will be large regardless of the ground truth class’s sensitivity. Our algorithm takes advantage of this behavior to promote robustness and maintain contextual sensitivity.

\section{Alternative Contrastive Loss Function}\label{appendix:loss}

As mentioned in the body of the paper, we can apply FCL with other contrastive losses. An alternative choice is the original version of the contrastive loss as proposed by \citep{Chopra2005}.
\begin{align}
\ell_{\text{margin}}^i & = \distance(z_i,z_i^{+})^2 + \max(0, \gamma - \distance(z_i,z_i^{-}))^2, \label{eq:fcl-margin}
\end{align}
Eq.~\eqref{eq:fcl-margin} and Eq.~\eqref{eq:fcl-xe} solve similar problems, taking different approaches. Eq.~\eqref{eq:fcl-margin} strictly minimizes the distance between $z_i$ and $z_i^{+}$ (to be zero at its optimal) and encourages a margin of at least $\gamma$ between $z_i$ and $z_i^{-}$. Eq~\eqref{eq:fcl-xe}, on the other hand, applies a softer contrast between the rankings of $\similarity(z_i, z_i^{+})$ and $\similarity(z_i, z_j \in \mathcal{N}_i)$ similar to the softmax cross entropy loss. Another difference is that Eq.~\eqref{eq:fcl-xe} uses a larger negative set and enforces cross example rankings. 
We explored this alternative loss~\eqref{eq:fcl-margin} in the synthetic MNIST experiment (\S\ref{sec:exp-synthetic}) and saw similar results to the original loss.

\section{Experimental Setup}\label{appendix:exp}

\paragraph{Architecture}
For CIFAR experiments, we used a ResNet-56 architecture, with the following configuration for each ResNet block $(\text{n}_\text{layer}, \text{n}_\text{filter}, \text{stride})$: [(9, 16, 1), (9, 32, 2), (9, 64, 2)].

For ImageNet experiments, we used a ResNet-v2-50 architecture, with the following configuration for the ResNet block  $(\text{n}_\text{layer}, \text{n}_\text{filter}, \text{stride})$: [(3, 64, 1), (4, 128, 2), (6, 256, 2), (3, 512, 2)].

\paragraph{Optimization}
For CIFAR, we used SGD momentum optimizer (Nesterov=True, momentum=0.9) with a linear learning rate ramp up for 15 epochs (peaked at 1.0) and a step-wise decay of factor 10 at epochs 200, 300, and 400. In total, we train for 450 epochs with a batch size of 1024.

For ImageNet, we also used SGD momentum optimizer (Nesterov=False, momentum=0.9) with a linear learning rate ramp up for the first 5 epochs (peaked at 0.8) and decayed by a factor of 10 at epochs 30, 60 and 80. In total, we train for 90 epochs with a batch size of 1024.

\paragraph{Hyperparameters}
We provide additional hyperparameter details for the experiments. (PG stands for Patch Gaussian):
\begin{itemize}
    \item \textbf{SST}\\
    FCL $k_{\text{ratio}}=0.1, \tau=0.02, \lambda=0.5$
    \item \textbf{MNIST}\\
    FCL $\sigma=0.5, \tau=0.1, \lambda=0.001$
    \item \textbf{Noisy CIFAR-10}\\
    XE+Gaussian $\sigma=0.3, \lambda=0.0001$\\
    CL+Gaussian $\sigma=0.5, \tau=0.5, \lambda=0.0001$, ramp\_up=14000steps\\
    FCL $k=256, \sigma=0.5, \tau=2, \lambda=0.0001$, ramp\_up=14000steps\\
    PG $\sigma=0.1$, patch\_size=25\\
    PG+FCL $k=256, \sigma=0.5, \tau=1, \lambda=0.0001$, ramp\_up=14000steps (PG $\sigma=0.1$, patch\_size=25)
    \item \textbf{Noisy CIFAR-100}\\
    XE+Gaussian $\sigma=0.3, \lambda=0.0001$\\
    CL+Gaussian $\sigma=0.5, \tau=0.5, \lambda=0.0001$, ramp\_up=10000steps\\
    FCL $k=256, \sigma=0.5, \tau=0.1, \lambda=0.0001$, ramp\_up=10000steps\\
    PG $\sigma=0.05$, patch\_size=25\\
    PG+FCL $k=256, \sigma=0.5, \tau=0.1, \lambda=0.0001$, ramp\_up=10000steps (PG $\sigma=0.05$, patch\_size=25)
    \item \textbf{CIFAR-10-C}\\
    XE+Gaussian $\sigma=0.3, \lambda=0.0001$\\
    CL+Gaussian $\sigma=0.5, \tau=0.5, \lambda=0.0001$, ramp\_up=14000steps\\
    FCL $k=256, \sigma=0.5, \tau=2, \lambda=0.0001$, ramp\_up=14000steps\\
    PG $\sigma=0.1$, patch\_size=25\\
    PG+FCL $k=256, \sigma=0.5, \tau=1, \lambda=0.0001$, ramp\_up=10000steps (PG $\sigma=0.1$, patch\_size=25)
    \item \textbf{CIFAR-100-C}\\
    XE+Gaussian $\sigma=0.3, \lambda=0.0001$\\
    CL+Gaussian $\sigma=0.5, \tau=0.5, \lambda=0.0001$, ramp\_up=10000steps\\
    FCL $k=256, \sigma=0.5, \tau=0.1, \lambda=0.0001$, ramp\_up=10000steps\\
    PG $\sigma=0.1$, patch\_size=25\\
    PG+FCL $k=256, \sigma=0.5, \tau=0.1, \lambda=0.0001$, ramp\_up=10000steps (PG $\sigma=0.05$, patch\_size=25)
    \item \textbf{ImageNet-C}\\
    XE+Gaussian $\sigma=0.5, \lambda=0.0001$\\
    CL+Gaussian $\sigma=0.5, \tau=0.5, \lambda=0.0001$, ramp\_up=78000steps\\
    FCL $k=512, \sigma=1.0, \tau=0.5, \lambda = 0.0002$\\
    PG $\sigma=1.0$, patch\_size $\le$ 250\\
    PG+FCL $k=2048, \sigma=0.5, \tau=1.0, \lambda=0.0004$, ramp\_up=78000steps (PG $\sigma=1.0$, patch\_size $\ge$ 250)
\end{itemize}

\clearpage
\section{Full CIFAR-10-C, CIFAR-100-C and ImageNet-C Accuracy}\label{appendix:full_corruptions_accuracy}

\begin{table*}[!h]
    \footnotesize
    \begin{tabular}{@{}llrrrrrrr@{}}
        \toprule
        Dataset & Method 
        & \multicolumn{3}{c}{Noise} 
        & \multicolumn{4}{c}{Blur} \\
        & & gauss. & shot & impulse 
        & defocus & glass & motion & zoom\\
        \toprule
        CIFAR-10-C 
        & XE
        & 0.4049 & 0.5374 & 0.5477
        & 0.7912 & 0.4846 & 0.7350 & 0.7222\\
        & XE+Gaussian
        & 0.5029 & 0.6221 & 0.6149
        & 0.7992 & 0.5192 & 0.7315 & 0.7369\\
        & CL+Gaussian
        & 0.4767 & 0.5922 & 0.5650
        & 0.7919 & 0.5383 & 0.7238 & 0.7215 \\
        & FCL
        & 0.5505 & 0.6431 & 0.7311
        & 0.7922 & 0.5140 & 0.7273 & 0.7210\\
        \cmidrule{2-9}
        & PG+XE
        & \bf 0.8995 & \bf 0.9082 & 0.8776 
        & 0.8252 & 0.6731 & \bf 0.7634 & 0.7883\\
        & PG+FCL
        & 0.8983 & 0.9078 & \bf 0.8918
        & \bf 0.8268 & \bf 0.6764 & 0.7601 & \bf 0.7924\\
        \toprule
        CIFAR-100-C
        & XE 
        & 0.1644 & 0.2446 & 0.2249
        & 0.5577 & 0.2022 & 0.4884 & 0.4808\\
        & XE+Gaussian
        & 0.2024 & 0.2816 & 0.2667
        & 0.5557 & 0.2055 & 0.4812 & 0.4807\\
        & CL+Gaussian
        & 0.1880 & 0.2668 & 0.2502
        & 0.5562 & 0.2110 & \bf 0.4903 & 0.4827\\
        & FCL
        & 0.2551 & 0.3186 & 0.4847
        & 0.5571 & 0.2137 & 0.4861 & 0.4852\\
        \cmidrule{2-9}
        & PG+XE
        & 0.1773 & 0.2542 & 0.2279
        & \bf 0.5596 & 0.2001 & 0.4897 & \bf 0.4883\\
        & PG+FCL
        & \bf 0.2729 & \bf 0.3349 & \bf 0.5020 
        & 0.5487 & \bf 0.2310 & 0.4845 & 0.4768\\
        \toprule
        ImageNet-C
        & XE
        & 0.2860 & 0.2651 & 0.2335
        & 0.2945 & 0.2312 & 0.2844 & 0.3163\\
        & XE+Gaussian
        & 0.2876 & 0.2654 & 0.2339
        & 0.2978 & 0.2293 & 0.2851 & 0.3193\\
        & CL+Gaussian
        & 0.2898 & 0.2694 & 0.2383
        & 0.2955 & 0.2290 & 0.2873 & 0.3177\\
        & FCL
        & 0.2954 & 0.2738 & 0.2395 
        & 0.2989 & 0.2374 & 0.2861 & 0.3176\\
        \cmidrule{2-9}
        & PG+XE 
        & 0.3265 & 0.3070 & \bf 0.2822 
        & 0.3333 & \bf 0.2571 & \bf 0.2923 & \bf 0.3337\\
        & PG+FCL 
        & \bf 0.3304 & \bf 0.3107 & 0.2821
        & \bf 0.3344 & \bf 0.2571 & 0.2920 & 0.3302\\
        \bottomrule
    \end{tabular}
    \begin{tabular}{@{}llrrrrrrrr@{}}
        \toprule
        Dataset & Method 
        & \multicolumn{4}{c}{Weather} 
        & \multicolumn{4}{c}{Digital} \\
        & & snow & forest & fog & bright 
        & contrast & elastic & pixel & JPEG\\
        \toprule
        CIFAR
        & XE
        & 0.7933 & 0.7486 & 0.8587 & \bf 0.9231
        & \bf 0.7310 & 0.8092 & 0.6991 & 0.7756\\
        -10-C & XE+Gaussian
        & 0.8027 & 0.7711 & \bf 0.8601 & 0.9218
        & 0.7240 & 0.8104 & 0.7195 & 0.7750\\
        & CL+Gaussian
        & 0.7930 & 0.7589 & 0.8542 & 0.9189
        & 0.7267 & 0.8023 & 0.7083 & 0.7747\\
        & FCL
        & 0.7988 & 0.7555 & 0.8590 & 0.9217
        & 0.7233 & 0.8092 & 0.7036 & 0.7758\\
        \cmidrule{2-10}
        & PG+XE
        & 0.8306 & 0.8353 & 0.8303 & 0.9198 
        & 0.7085 & \bf 0.8417 & 0.7919 & \bf 0.8664\\
        & PG+FCL
        & \bf 0.8324 & \bf 0.8374 & 0.8256 & 0.9189
        & 0.6937 & 0.8416 & \bf 0.7994 & 0.8652\\
        \toprule
        CIFAR
        & XE
        & 0.5023 & 0.4370 & \bf 0.5874 & \bf 0.6842
        & \bf 0.4884 & 0.5458 & 0.4500 & 0.4578\\
        -100-C & XE+Gaussian
        & \bf 0.5061 & \bf 0.4403 & 0.5825 & 0.6805
        & 0.4793 & 0.5431 & 0.4582 & 0.4585\\
        & CL+Gaussian
        & 0.5034 & 0.4392 & 0.5822 & 0.6807
        & 0.4796 & \bf 0.5484 & 0.4560 & 0.4621\\
        & FCL
        & 0.5007 & 0.4333 & 0.5748 & 0.6779
        & 0.4640 & 0.5441 & 0.4583 & \bf 0.4724\\
        \cmidrule{2-10}
        & PG+XE
        & 0.4970 & 0.4308 & 0.5834 & 0.6819
        & 0.4843 & 0.5483 & \bf 0.4635 & 0.4623\\
        & PG+FCL
        & 0.5050 & \bf 0.4403 & 0.5756 & 0.6750
        & 0.4687 & 0.5394 & 0.4634 & 0.4688\\
        \toprule
        ImageNet 
        & XE
        & 0.2773 & 0.3304 & 0.4695 & 0.6083 
        & 0.3273 & 0.4096 & 0.2998 & 0.4763\\
        -C & XE+Gaussian
        & 0.2748 & 0.3323 & 0.4736 & 0.6088
        & 0.3298 & 0.4101 & 0.2989 & 0.4743\\
        & CL+Gaussian
        & 0.2745 & 0.3327 & 0.4728 & 0.6091
        & 0.3309 & 0.4060 & 0.2978 & 0.4768\\
        & FCL
        & 0.2739 & 0.3297 & 0.4674 & 0.6044
        & 0.3278 & 0.4143 & 0.3111 & 0.4777\\
        \cmidrule{2-10}
        & PG+XE 
        & \bf 0.2891 & 0.3464 & 0.4735 & \bf 0.6110
        & 0.3352 & \bf 0.4331 & 0.3232 & \bf 0.4939\\
        & PG+FCL 
        & 0.2880 & \bf 0.3495 & \bf 0.4761 & 0.6097
        & \bf 0.3368 & 0.4313 & \bf 0.3290 & 0.4934\\
        \bottomrule
    \end{tabular}
    \caption{Image classification accuracies on the CIFAR-10-C, CIFAR-100-C and ImageNet-C sets \citep{hendrycks2019benchmarking}. PG stands for Patch Gaussian data augmentation \citep{lopes2019improving}. All numbers are averaged by 5 runs.}
\end{table*}

\end{document}